\documentclass[12pt]{article}
\usepackage{amssymb,amsmath,cite,bm}
\usepackage[dvips]{graphicx,color}
\usepackage{psfrag,subfigure}
\newtheorem{proposition}{Proposition}

\newcommand{\norm}[1]{\left\Vert#1\right\Vert}

\newcommand{\diag}[1]{{\rm diag}\left(#1 \right)}

\title{\bf Adaptive Machine Learning for Cooperative Manipulators}

\author{Farhad Aghili\thanks{email: faghili@encs.concordia.ca}}

\date{}

\begin{document}

\maketitle

\begin{abstract}
The problem of self-tuning control of cooperative manipulators forming a closed kinematic chain in the presence of an inaccurate kinematics model is addressed using adaptive machine learning. The kinematic parameters pertaining to the relative position/orientation uncertainties of the interconnected manipulators are updated online by two cascaded estimators in order to tune a cooperative controller for achieving accurate motion tracking with minimum-norm actuation force. This technique permits accurate calibration of the relative kinematics of the involved manipulators without needing high precision end-point sensing or force measurements, and hence it is economically justified. Investigating the stability of the entire real-time estimator/controller system reveals that the convergence and stability of the adaptive control process can be ensured if $i)$ the direction of the angular velocity vector does not remain constant over time, and $ii)$ the initial kinematic parameter error is upper bounded by a scaler function of some known parameters. The adaptive controller is proved to be singularity-free even though the control law involves inverting the approximation of a matrix computed at the estimated parameters. Experimental results demonstrate the sensitivity of the tracking performance of the conventional inverse dynamic control scheme to kinematic inaccuracies, while the tracking error is significantly reduced by the self-tuning cooperative controller.
\end{abstract}

%\cite{Aghili-2012a}
%\cite{Aghili-2011b}  
%\cite{Prevot-Gourdeau-Aghili-Piedbouf-2004}
%\cite{Aghili-Nmavar-2006}
%\cite{Aghili-2009b}
%\cite{Aghili-2019a}
%\cite{Aghili-2019b}
%\cite{Aghili-2019c}
%\cite{Aghili-2017a}
%\cite{Aghili-2015b} 
%\cite{Aghili-2015c} 
%\cite{Aghili-2010i} 
%\cite{Aghili-2010h} 
%\cite{Aghili-2010}
%\cite{Aghili-2010c} 
%\cite{Aghili-Parsa-2009b}
%\cite{Aghili-2009c} 
%\cite{Aghili-2009g}
%\cite{Aghili-2008e} 
%\cite{Aghili-2005}  
%\cite{Namvar-Aghili-2005}
%\cite{Namvar-Aghili-2004a}   
%\cite{Aghili-Piedboeuf-2003a} 
%\cite{Aghili-2003} 

%------------------------------------------------------
\section{Introduction}
%------------------------------------------------------
In some industrial applications, it is necessary to move a large and
heavy payload, such as a vehicle chases, by two or more manipulators
to share the load and also to gain stiffness. Cooperative
multi-robot systems form a closed kinematic chain by each robot
end-effector grasping a location on the object \cite{Aghili-2012a,Prevot-Gourdeau-Aghili-Piedbouf-2004}. This configuration
literally combines the cooperative serial robots into a single
redundant parallel mechanism with a high structural stiffness and
large lifting capacity. However, when cooperative manipulators pick up objects of different dimensions, i.e., uncertain orientation and length at griping points, the overall kinematics of the interconnected robotic system changes and therefore the kinematic model is required to be updated unless the relative position/orientation of the manipulators are precisely known. However, to calibrate  the relative kinematic parameters  of the manipulators requires  precision endpoint or camera-based measurement systems~\cite{Park-Lee-Cho-2011}. This work proposes a self-tuning control of cooperative manipulators  to track a motion trajectory, without knowing the true kinematic parameters, and at the same time identify the parameters.

The underlying control issue of how to synchronously move the involved end-effectors of a cooperative manipulator system to track a predefined motion trajectory and at the same time regulate the  internal forces has been the focus of many research works~\cite{Alford-Belyen-1984,Tarn-Bejczy-Yuan-1986,Aghili-Parsa-2009b,Tao-Luh-Zheng-1987,Hayati-1986,Nakamura-Nagai-Yoshikawa-1989,Aghili-2019c,Yao-Gao-Chan-1992,Arimoto-Miyazaki-Kawamura-1993,Pagilla-Tomizuka-1994}.
However, one of the remaining problem in synchronous control of
cooperative manipulators is the sensitivity of the interconnected
robotic system to the uncertainties of the closed-kinematic loop.
Even small kinematic errors can give rise to significant control
error and/or actuation force unless an impedance control or force
feedback is used to relieve the build-up internal forces.
Although the forward kinematics of commercial manipulators are usually known, the relative position/orientation of manipulators  which are set up to perform cooperative manipulation tasks may not be precisely known. That  introduces modeling uncertainty in the overall closed kinematic chain and therefore precise and effective cooperative control of the manipulators may require closed-loop kinematic calibration of one sort or another.

There are various  non-adaptive and adaptive methods for control of multiple cooperative
manipulators. The earliest approach was based on  master/slave
configuration of multiple robots
\cite{Alford-Belyen-1984,Tarn-Bejczy-Yuan-1986,Tao-Luh-Zheng-1987}
in which the master robot is position controlled while and slave
robots are under force control to maintain kinematic constraints.
Another approach developed later considers the cooperative manipulators and the grasped object as a
closed kinematic chain~\cite{Hayati-1986,Nakamura-Nagai-Yoshikawa-1989,Aghili-Nmavar-2006,Yao-Gao-Chan-1992,Namvar-Aghili-2005,Arimoto-Miyazaki-Kawamura-1993,Aghili-2010,Pagilla-Tomizuka-1994,Namvar-Aghili-2004a}. Then, a hybrid position/force control
strategy is utilized to simultaneously
control both the pose of the object and the internal forces\cite{Raibert-Craig-1981}. An
internal force-based impedance control scheme for cooperating
manipulators is introduced  in \cite{Bonitz-Hsia-1996,Aghili-2015c,Caccavale-Chiacchio-Mario-Villani-2008,Aghili-2010i}. The
controller uses the forces sensed at the robot end effectors to
compensate for the effects of the objects' dynamics. Active
compliance is employed in \cite{Arai-Osumi-Fukuoka-1995} in order to
accommodate positioning errors between the double cooperative
manipulators. Adaptive cooperative control of manipulators with uncertain
inertial parameters has been also investigated in the
literature
\cite{Walker-Kim-Dionise-1989,Hu-Goldenberg-1989,Namvar-Aghili-2005,Zribi-Ahmed-1991,Aghili-2010h,Sun-Mills-2002,Aghili-2010c,Jean-Fu-1993,Aghili-2008e,Gueaieb-Karray-Al-Sharhan-2007}.

Walker {\em et al.} proposed an adaptive cooperative control scheme
for manipulators carrying an object with unknown parameters.
Adaptive control of cooperative manipulators with an uncertain dynamic
model to achieve asymptotic convergence of both position tracking
errors and force tracking errors was presented in
\cite{Hu-Goldenberg-1989}. Robust adaptive motion control of
multiple robots in the face of bounded disturbance was addressed in
\cite{Zribi-Ahmed-1991}. A decentralized adaptive control scheme
where each robot is controlled separately by its own local
controller is presented in \cite{Sun-Mills-2002}. The adaptive
coordinated control of multiple robots transporting an object where
all dynamics parameters of the robots and object are unknown
constants is further developed in \cite{Jean-Fu-1993}. Adaptive
hybrid force/motion control of cooperative manipulators interacting
with unknown environment is proposed in \cite{Namvar-Aghili-2005,Aghili-2009g}.
Although these adaptive  cooperated control schemes can deal with modeling
uncertainties in dynamics of the robots and object, the kinematics of interconnected robotic system is assumed to
be exactly known. On the other hand, experiments have shown that even small kinematic inaccuracy, due to tolerance and geometric uncertainties, can significantly deteriorate the tracking performance. This work is  motivated by development of a self-tuning adaptive control of cooperative manipulators that can deal with the kinematic inaccuracy of the interconnected robotic system. The main advantage of the cooperative control approach is that it does not use force measurements to accommodate the position/oritntation errors between the cooperative manipulators. Rather the controller is able to tune itself adaptively by identifying the parameters contributing to the errors. Moreover, since no external positioning device is required, such as a high precision camera system, the control approach is advantageous for calibrating the relative pose of cooperative manipulators. However, it should be pointed out that the control method works only if the common object is a rigid-body and that the stability can be ensured upon certain conditions. The contributions of the this work are the development of an estimator for the kinematic parameters and rigorous analysis of the entire estimator/control system leading to derivation of the stability conditions.

This paper presents an adaptive machine learning approach for control of cooperative manipulators carrying a common
object in the presence of geometric uncertainties~\cite{Aghili-2012a,Aghili-2011b}. The self-tuning control method allows precise calibration of the relative kinematic parameters of the manipulators  while performing a cooperative manipulation task. The unknown position/orientation variables of the
closed-chain loop are estimated in real-time to tune a cooperated
controller in order to achieve motion tracking of a reference
trajectory while minimizing the Euclidean norm of the actuator
forces. It is systematically proved that under the conditions
that the direction of vector the angular velocity of the object
changes over time and that initial kinematic error is sufficiently small, the convergence and stability of the combined
on-line estimator and controller can be ensured. Finally, comparative
experimental results demonstrating the tracking performance of the
adaptive controller are  presented.

%Fig. 1=============================================================
\begin{figure}[t]
\centering
\includegraphics[width=9cm]{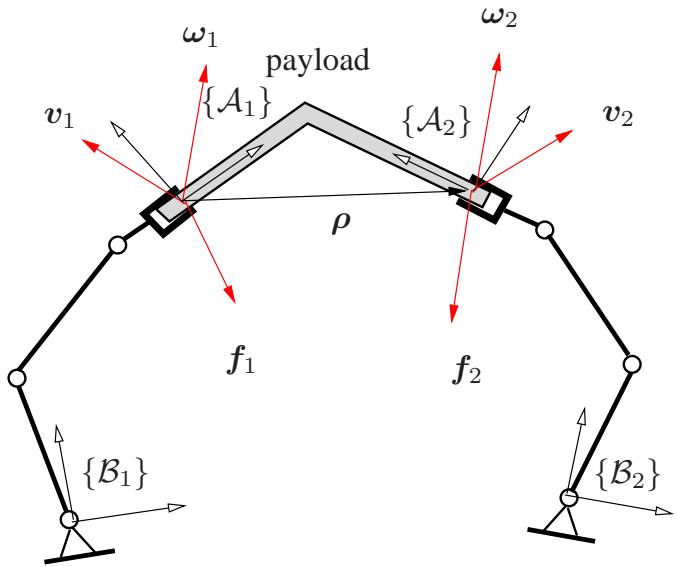} \caption{Cooperative robots carrying a common object.} \label{fig:coop_robot}
\end{figure}
%=============================================================

\section{Adaptive Cooperation Control with Kinematics Uncertainties}

\subsection{On-line Parameter Estimation}
Fig.\ref{fig:coop_robot} illustrates two cooperative manipulators
carrying a common object. Coordinate frames $\{ {\cal A}_i \}$  and $\{ {\cal B}_i \}$ are attached to the base and end-effector of the $i$th manipulator, respectively. Here, we make the following fundamental assumptions on the kinematics model of the interconnected robotic system
\begin{enumerate}
\item The relative position/orientation of the end-effectors grasping on to  the object, i.e., $\{ {\cal A}_1 \}$  w.r.t. $\{ {\cal A}_2 \}$,  are not precisely known.
\item The relative position/orientation of the manipulators' bases, i.e.,  $\{ {\cal B}_1 \}$  w.r.t. $\{ {\cal B}_2 \}$, are not precisely known.
 \end{enumerate}
However, all manipulators forward kinematics are supposed to be known.
Assume that $\bm v_j$ and $\bm\omega_j$ be the linear and angular
velocities of the $j$th robot's end-effector expressed in the tool
frame $\{ {\cal A}_j \}$. Notice that $\bm\omega_1$ and $\bm\omega_2$ are the same vector expressed in different coordinate frames.  The end-effector velocities can be
obtained from the kinematics model of the manipulators and the measurements of the manipulators' joint angles and joint velocities as:
\begin{equation}
\begin{bmatrix} \bm v_j \\ \bm\omega_j \end{bmatrix}= \bm J_j(\bm q_j) \dot{\bm
q}_j \qquad j=1,2.
\end{equation}
where $\bm J_j$ is the Jacobian matrix of the $j$th manipulator
expressed in the tool frame\footnote{The Jacobian expressed in the tool frame $^{{\cal A}}\bm J$ is related to Jacobian expressed in the base frame $^{{\cal B}}\bm J$ by  \[ ^{{\cal A}}\bm J = \begin{bmatrix} \bm R^T & \bm 0 \\ \bm 0 & \bm R^T \end{bmatrix}  {^{{\cal B}}\bm J}, \] where $\bm R$ is the rotation matrix representing the end-effector orientation.  Here, $^{{\cal A}}\bm J=\bm J$ for simplicity of notation.} $\{ {\cal A}_j \}$ and $\bm q_j$
represent the joint angles of the corresponding manipulator. Also
assume that the position and orientation of the end-effectors
grasping the object are represented by vector $\bm\rho$, which is
expressed in $\{ {\cal A}_2 \}$,  and rotation matrix $\bm A$, which
represents the orientation of coordinate frame $\{ {\cal A}_2 \}$
w.r.t. $\{ {\cal A}_1 \}$. The orientation can be also represented
by the unit quaternion $\bm\eta$, which is related to the rotation
matrix by
\begin{equation} \label{eq:R}
\bm A(\bm\eta) = (2 \eta_o^2-1) \bm I + 2 \eta_o [ \bm\eta_v \times]
+ 2 \bm\eta_v \bm\eta_v^T.
\end{equation}
Here, $\bm I$ is the identity matrix with adequate dimension,
$\bm\eta_v$ is the vector part of the quaternion and $\eta_o$ is the
scalar part, i.e., $\bm\eta=[\bm\eta_v^T \; \eta_o]^T$. Then, due to
the closed-loop kinematics, the velocities of the end-effectors are
related to one other by
\begin{subequations} \label{eq:trans}
\begin{align} \label{eq:omega_2}
\bm\omega_2 &= \bm A(\bm\eta) \bm\omega_1  ,\\ \label{eq:v_2} \bm
v_2 & = \bm A(\bm\eta) \bm v_1 - \bm\rho \times \bm\omega_2.
\end{align}
\end{subequations}
Our estimation strategy is to solve \eqref{eq:trans} for parameters
$\bm\rho$ and $\bm\eta$ such that the residual errors are minimized
in the least-squares sense. That is, given the measurements of the
linear and angular velocities $\bm v_{1_i}$, $\bm v_{2_i}$,
$\bm$$\bm\omega_{1_i}$, and $\bm\omega_{2_i}$ at time $t_i$, we have
\begin{equation} \label{eq:two_epsilon}
\begin{split}
\min_{\bm\eta, \bm\rho} &  \; \frac{1}{2}\sum_{i=1}^k a_i \left\| \begin{bmatrix}\bm\epsilon_{\omega_i} \\ \bm\epsilon_{v_i} \end{bmatrix} \right\|^2 \\
{\rm s.t.} \; &\|\bm\eta \| =1
\end{split}
\end{equation}
where
\begin{subequations}
\begin{align} \label{eq:epsilon_w}
\bm\epsilon_{\omega_i}(\bm\eta)& =\bm\omega_{2_i} -\bm A(\bm\eta)
\bm\omega_{1_i} \\ \label{eq:epsilon_v} \bm\epsilon_{v_i}(\bm\eta,
\bm\rho) & =  \bm v_{2_i} - \bm A(\bm\eta) \bm v_{1_i} + \bm\rho
\times
 \bm\omega_{2_i}
\end{align}
\end{subequations}
are the residual errors, and $a_i$ are  non-negative weights, which
can be selected according to the time varying forgetting function
\begin{equation} \label{eq:forgetting}
a_i= e^{-\mu(t_k - t_i)}
\end{equation}
with $\mu<1$ being a fading memory factor. Problem
\eqref{eq:two_epsilon} is difficult to simultaneously solve for
$\bm\rho$ and $\bm\eta$. However, one may notice that the residual
error $\bm\epsilon_{\omega_i}$ is a function of only $\bm\eta$. This
may suggest that estimations of $\bm\eta$ and $\bm\rho$ can be
sequentially obtained as follow: Firstly, find quaternion estimate $\hat{\bm\eta}$ by minimizing the residual error in
\eqref{eq:epsilon_w}. Then, use the quaternion estimate to obtain displacement vector estimate
$\hat{\bm\rho}$ by minimizing the second residual error in \eqref{eq:epsilon_v}.

Finding the orientation from the vector observation
\eqref{eq:omega_2} is tantamount to the Wahba's problem
\cite{Wahba-1965} that minimizes the cost function
\begin{equation} \label{eq:J1}
J_{\omega}(t_k) = \frac{1}{2} \sum_{i=1}^k a_i \| \bm\epsilon_{\omega_i} \|^2 ,
\end{equation}
There are many algorithm to solve the above optimization problem
\cite{Lerner-1978,Shuster-Oh-1981,Markley-Mortari-2000,Crassidid-Markley-Cheng-2007},
of which the parametrization of the orientation by a unit quaternion
is the most elegant and useful one. It becomes immediately clear from expanding the expression on RHS of \eqref{eq:J1} in terms of the angular velocities,
\begin{equation} \label{eq:J_epxpansion}
J_{\omega} = \sum_{i=1}^k a_i \big( \| \bm\omega_{1_i} \|^2 +  \| \bm\omega_{2_i} \|^2 \big) -  \sum_{i=1}^k a_i \bm\omega_{2_i}^T \bm A \bm\omega_{1_i},
\end{equation}
that the cost function is minimized when the last term in RHS of \eqref{eq:J_epxpansion} is maximized. Moreover, using identity \eqref{eq:R}, one can show that
\begin{equation}
\bm\omega_{2_i}^T \bm A \bm\omega_{1_i} = \bm\eta^T \bm\Omega(\bm\omega_{1_i},\bm\omega_{2_i}) \bm\eta,
\end{equation}
where
\begin{equation} \notag
\bm\Omega(\bm\omega_1,\bm\omega_2) = \begin{bmatrix} \bm\omega_{2}\bm\omega_{1}^T + \bm\omega_{1}\bm\omega_{2}^T - \bm\omega_{1}^T\bm\omega_{2} \bm I & \bm\omega_{2} \times \bm\omega_{1}  \\ (\bm\omega_{2} \times \bm\omega_{1})^T & \bm\omega_{1}^T \bm\omega_{2} \end{bmatrix}.
\end{equation}
Therefore, finding the optimal quaternion minimizing cost function \eqref{eq:J1} is tantamount to solving the following
quadratic programming
\begin{align} \label{eq:quest}
&\max \hat{\bm\eta}_k^T \bm\Gamma_k \hat{\bm\eta}_k \\ \notag  {\rm s.t.} & \quad
\| \hat{\bm\eta}_k \|^2 - 1 =0,
\end{align}
where $\hat{\bm\eta}_k$ is the optimal quaternion estimate at time $t_k$, and
\begin{equation} \label{eq:summation_Gamma}
\bm\Gamma_k = \sum_{i=1}^k a_i \bm\Omega(\bm\omega_{1_i},\bm\omega_{2_i}).
\end{equation}
Denote the Lagrangian equation of the above constrained quadratic programming as
\begin{equation} \notag
{\cal L}=\bm\eta^T \bm\Gamma_k \bm\eta - \lambda(\bm\eta^T \bm\eta -1 ),
\end{equation}
where $\lambda$ is the Lagrangian multiplier. Then, the optimal solution has to satisfy the stationary condition, which leads to the following equation:
\begin{equation} \label{eq:eta_est}
\frac{\partial {\cal L}}{\partial \bm\eta} =0 \quad \Rightarrow \quad \bm\Gamma_k \hat{\bm\eta}_k - \lambda_{\rm max} \hat{\bm\eta}_k = \bm 0.
\end{equation}
Notice that \eqref{eq:eta_est} is equivalent to the characteristic equation of $\bm\Gamma_k$
for the largest eigenvalue $\lambda_{\rm max}=\lambda_{\rm max}(\bm\Gamma_k)$. In other words, the optimal quaternion is the eigenvector of matrix  $\bm\Gamma_k$ corresponding to its largest eigenvalue.

One can readily show that instead of summation
\eqref{eq:summation_Gamma}, matrix $\bm\Gamma_k$ can be updated
recursively as
\begin{equation} \label{eq:recursive_Gamma}
\bm\Gamma_k = \varrho \bm\Gamma_{k-1} +\bm\Omega(\bm\omega_{1_k},\bm\omega_{2_k}),
\end{equation}
where
\begin{equation} \label{eq:varrho}
\varrho = e^{-\mu h},
\end{equation}
and $h$ is the sample interval. Despite the fact that \eqref{eq:recursive_Gamma}
and \eqref{eq:summation_Gamma} are equivalent, the recursive formula
\eqref{eq:recursive_Gamma} is advantageous because of its
computational efficiency and admitting  initial
quaternion  $\bm\eta_0$, where $\bm\eta_0^T \bm\eta_0=1$. By
inspection, one can show that initial matrix
\begin{equation} \notag
\bm\Gamma_0 = \bm\eta_0 \bm\eta_0^T,
\end{equation}
with $\lambda_{\rm max}=1$, satisfies the eigenvalue equation
\eqref{eq:eta_est}.

In the absence of measurement noise, the optimal solution zeros the
cost function \eqref{eq:J1}. However, the estimated quaternion does
not necessarily converge to the actual value unless the velocity
signals satisfy the {\em persistent excitation} condition.

\begin{proposition} \label{prop:eta}
If the direction of the angular velocity does not remain constant
over interval  $[t_k \quad t_{k+p}]$, then the estimator
\eqref{eq:eta_est} is convergent, i.e., $\hat{\bm\eta}$ converges to $\bm\eta$.
\end{proposition}
{\sc Proof:} Assume that the direction of the object angular
velocity vector changes  at two instances of time, $t_m$ and $t_n$,
within the time interval. Then, denoting $\hat{\bm A} = \bm A(\hat{\bm\eta})$, we can say
\begin{subequations} \label{eq:omegaA_hat}
\begin{align}
\bm\omega_{2_m} - \hat{\bm A} \bm\omega_{1_m}&= \bm 0\\
\bm\omega_{2_n} - \hat{\bm A} \bm\omega_{1_n}&= \bm 0,
\end{align}
\end{subequations}
which can be written as
\begin{align*}
\tilde{\bm A} \bm\omega_{2_m} - \bm\omega_{2_m} &= \bm 0\\
\tilde{\bm A} \bm\omega_{2_n} - \bm\omega_{2_n} &= \bm 0,
\end{align*}
where $\tilde{\bm A} = \hat{\bm A} \bm A^T$ is the rotation matrix of the estimation error. The above equations are equivalent to the characteristic equation of matrix $\tilde{\bm A}$ for two linearly independent eigenvectors, $\bm\omega_{2_m} \nparallel \bm\omega_{2_n}$, with eigenvalue $1$. Moreover, since the products of all  eigenvalues of a rotation matrix is one, i.e.,
\begin{equation} \notag
\prod_{i=1}^3 \lambda_i(\tilde{\bm A})= \det(\tilde{\bm A})=1
\end{equation}
the third eigenvalue of $\tilde{\bm A}$ must be $1$ too. Consequently, the only possibility for the rotation matrix $\tilde{\bm A}$ with all of eigenvalues equal to one is the identity matrix  meaning that $\hat{\bm
A} = \bm A$.

Now with the orientation estimate in hand, one can obtain an
estimate of the displacement vector $\bm\rho$ by making use of
\eqref{eq:epsilon_v}. To this end, \eqref{eq:v_2} is rewritten in
the form which is linear in terms of the unknown parameter
$\bm\rho$, i.e.,
\begin{equation}
[\bm\omega_{2_i} \times] \bm\rho = \bm
A(\hat{\bm\eta_i}) \bm v_{1_i} -\bm v_{2_i}.
\end{equation}
Analogous to \eqref{eq:J1}, the problem of finding the displacement
vector $\bm\rho$ is formulated as minimizing the prediction error
\begin{equation} \label{eq:J2}
J_v(t_k) = \frac{1}{2}\sum_{i=1}^k a_i \| \bm\epsilon_{v_i} \| ^2,
\end{equation}
where $a_i$, as defined in \eqref{eq:forgetting}, incorporates
exponential forgetting of data. Then, the parameter update law
minimizing \eqref{eq:J2} is given by the following recursive
least-squares estimator \cite{Slotine-Li-1991}
\begin{equation}\label{eq:rho_est}
\hat{\bm\rho}_{k+1} = \hat{\bm\rho}_k + \bm K_k \big(\bm v_{2_k} - \bm A(\hat{\bm\eta}_k) \bm v_{1_k} -  \bm\omega_{2_k} \times \hat{\bm\rho}_k \big)
\end{equation}
where the estimator gain is updated according to
\begin{align} \notag
\bm K_k &= -\bm P_k [\bm\omega_{2_k} \times] \big(\varrho  \bm I -
[\bm\omega_{2_k} \times] \bm P_k [\bm\omega_{2_k} \times]
\big)^{-1}\\ \label{eq:lsq_gain}
\bm P_{k+1} &= \frac{1}{\varrho}\big( \bm I - \bm K_k
[\bm\omega_{2_k}
\times] \big) \bm P_k
\end{align}
and scalar $\varrho$ was defined in \eqref{eq:varrho}; see Appendix~\ref{apdx:least-squares}.

If regressor $[\bm\omega_2 \times]$ satisfies the {\em
persistently exciting} condition, then estimator \eqref{eq:rho_est}
is convergent. The persistent excitation condition dictates there
exists $p>1$ such that
\begin{equation} \label{eq:pe}
\bm\Pi=  \sum_{i=k}^{k+p} -[\bm\omega_{2_i} \times]^2
> 0 \qquad \forall k>0
\end{equation}

\begin{proposition} \label{prop:rho}
If the direction of the angular velocity vector  does not remain
constant over the time interval $[t_k \; t_{k+p}]$, then the
persistent excitation condition \eqref{eq:pe} is satisfied and
estimator \eqref{eq:rho_est} is convergent, i.e., $\hat{\bm\rho}\rightarrow \bm\rho$ as $t\rightarrow \infty$.
\end{proposition}
{\sc Proof:} Matrix $-[\bm\omega_{2_i} \times]^2 \geq 0$ is rank
deficient and hence  it is positive
semi-definite but not positive definite. This is because for
any vector $\bm\omega\in\mathbb{R}^3$ we have
\begin{equation} \notag
\mbox{eigenvalues of } -[\bm\omega \times]^2 := \{ 0, \; \|\bm\omega \|^2 \}.
\end{equation}
Although matrix $-[\bm\omega_{2_i} \times]^2$ is singular
at every instance of time, the persistent excitation condition
requires that its integral over an interval of length $[t_k, \;
t_{k+p}]$ be uniformly positive definite. Assuming that the
direction of the angular velocity vector changes from time $t_m$ to
$t_n$, one can rewrite summation \eqref{eq:pe} as
\begin{equation} \label{eq:PI}
\bm\Pi = \bm\Upsilon  - \sum_{\scriptsize{\begin{array}{l}{i = k}\\{i
\neq n , m}\end{array}}}^{k+p} [\bm\omega_{2_i} \times]^2,
\end{equation}
where
\begin{equation} \notag
\bm\Upsilon = -[\bm\omega_{2_m} \times]^2 - [\bm\omega_{2_n} \times]^2.
\end{equation}
The summation in the right-hand side of \eqref{eq:PI} is at least a
positive semi-definite matrix. Since adding a positive semi-definite
to a positive definite matrix always results in another positive
definite matrix, positive definiteness of matrix $\bm\Pi$ rests on
showing that $\bm\Upsilon$ is a full-rank matrix. In a proof by
contradiction, we show that $\bm\Upsilon$ must be a full-rank matrix
provided that $\bm\omega_{2_m}$ and $\bm\omega_{2_n}$ are not
collinear. Since $\bm\Upsilon$ is the sum of two positive  semi-definite matrices, the only possibility for $\bm\Upsilon$ being rank deficient is that there exists a non-zero vector $\bm v\neq \bm 0$ such that $\bm v^T \bm\Upsilon \bm v =0$. The latter equation in conjunction with the positive definiteness of $-[\bm\omega_{2_m} \times]^2$ and $- [\bm\omega_{2_n} \times]^2$ require that both identities
\begin{equation}
\bm v^T [\bm\omega_{2_m} \times]^2 \bm v =0 \quad \mbox{and} \quad \bm v^T [\bm\omega_{2_n} \times]^2 \bm v =0
\end{equation}
are satisfied. The former and latter equations imply that $\bm v$ must be parallel with vectors $\bm\omega_{2_m}$ and $\bm\omega_{2_n}$, respectively. However, the only possibility of these to happen is that $\bm v$ is the zero vector because $\bm\omega_{2_m}$ and $\bm\omega_{2_n}$ are linearly independent.
This means that matrix
$\bm\Upsilon$ is full-rank and therefore $\bm\Pi$ is a positive
definite matrix.

\subsection{Parametric Design of Coordination Control}
%Fig. 2=============================================================
\begin{figure}
\centering
\includegraphics[width=10cm]{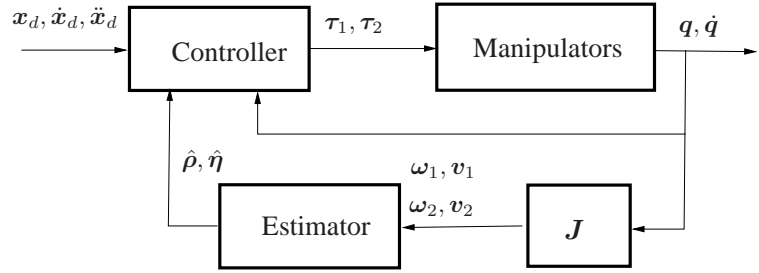} \caption{The sef-tuning cooperative control scheme.} \label{fig:controler_adaptive}
\end{figure}
%=============================================================

Fig.\ref{fig:controler_adaptive} schematically illustrates the
self-tuning cooperative controller  which incorporates the updated
kinematic parameters from the estimator. The concern is the stability of the entire estimation and control process and the tracking performance of the controller, given the time-varying kinematic parameters which starts from an initial guess at the parameters.

Assume that vectors $\bm x_i$ is the minimal representation of
the position and orientation of the $i$th robot end-effector. Then,
the following mapping is in order
\begin{equation} \label{eq:xL}
\dot{\bm x}_i \triangleq \bm L_i
\begin{bmatrix} \bm v_i \\ \bm\omega_i \end{bmatrix} \qquad \forall
i=1,2
\end{equation}
where $\bm L_i= \diag{\bm I , \bm L_o(\bm q_i)}$, and transformation
matrix $\bm L_o$ depends on a particular set of parameters used to
represent the orientation \cite{Canudas-Siciliano-Bastin-book-1996}.
Also, assume that the unknown kinematic parameters are placed in a single vector
\begin{equation} \notag
\bm\theta = \begin{bmatrix} \bm\rho \\ \bm\eta \end{bmatrix}.
\end{equation}
Then, in view of \eqref{eq:trans} and \eqref{eq:xL}, one can show that the generalized
velocities of the two manipulators are related by
\begin{equation} \label{eq:ddx2}
\dot{\bm x}_2 = \bm L_2 \bm T(\bm\theta) \bm L_1^{-1} \dot{\bm x}_1
\end{equation}
where
\begin{equation} \label{eq:T}
\bm T(\bm\theta) = \begin{bmatrix} \bm A(\bm\eta)^T & \bm A(\bm\eta)^T [\bm\rho \times] \\ \bm 0 & \bm A(\bm\eta)^T \end{bmatrix}
\end{equation}
is the velocity transformation matrix.  Time-derivative of \eqref{eq:ddx2} yields
\begin{equation}  \label{eq:ddotx2}
\ddot{\bm x}_2 = \bm L_2\bm T \bm L_1^{-1} \ddot{\bm x}_1 + \bm D \dot{\bm x}_1,
\end{equation}
where $\bm D= \dot{\bm L}_2 \bm T - \bm L_2 \bm T \bm L_1^{-1} \dot{\bm L}_1\bm L_1^{-1} $.

The dynamics of the $i$th manipulator in the task space
\cite{Canudas-Siciliano-Bastin-book-1996} can be concisely written
as
\begin{equation} \label{eq:dynamics_task}
\bm M_i \ddot{\bm x}_i + \bm h_i(\bm q_i, \dot{\bm q}_i) = \bm u_i -
\bm f_i, \qquad \forall i=1,2
\end{equation}
where
\begin{equation} \notag
\bm u_i = \bm J_i^{-T} \bm\tau_i,
\end{equation}
see Appendix~\ref{apdx:cartesian_dynamics}. Here, $\bm\tau_i$ is the vector of joint torques, $\bm f_i$ is
the vector of generalized force exerted at the end-effector,
$\bm M_i$ is the Cartesian mass matrix, and nonlinear vector $\bm h_i$
contains the Coriolis, centrifugal, and gravity terms. Assuming kinematic
singularity does not occur, there is a one-to-one correspondence
with the torque vector $\bm\tau_i$ and the corresponding auxiliary input $\bm u_i$ \cite{Aghili-2005,Aghili-Piedboeuf-2003a,Aghili-2003}.
Therefore, in the following derivation, we will take $\bm u_i$ as
the control input for the sake of simplicity. Moreover, writing the
balance of the forces on the common object yields
\begin{equation} \label{eq:dynmaics_obj}
\bm M_o \ddot{\bm x} + \bm h_o  = \bm f_1 + \bm T^T \bm f_2
\end{equation}

where $\bm x$ is the Cartesian coordinate of the object, $\bm M_o$ and $\bm h_o$ are the object inertia matrix and the
associated Coriolis, centrifugal, and gravity terms, and $\bm T^T$ is the corresponding force transformation. Define the augmented control input and generalized coordinates, respectively, as
\begin{equation} \notag
\bm u= \begin{bmatrix}\bm u_1 \\ \bm u_2 \end{bmatrix} \quad  \mbox{and} \quad  \bm q=\begin{bmatrix} \bm q_1 \\ \bm q_2\end{bmatrix}.
\end{equation}
The velocity of the object can be related to one of the involved end-effectors, say the first one, by
$\dot{\bm x}_1 = \bm\Lambda \dot{\bm x}$, where constant matrix $\bm\Lambda$ represents the corresponding velocity transformation. Thus, upon
substitution of $\bm f_1$, $\bm f_2$, and $\ddot{\bm x}_2$  from \eqref{eq:ddotx2} and \eqref{eq:dynamics_task} into\eqref{eq:dynmaics_obj}, the latter equation can  be arranged in the following form
\begin{equation} \label{eq:dyamics_uo}
\bar{\bm M} \ddot{\bm x} + \bar{\bm h} = \bar{\bm u}
\end{equation}
where
\begin{subequations}
\begin{align} \notag
\bar{\bm M} (\bm q , \bm\theta) & \triangleq \bm M_o + \big(\bm M_1 +
\bm T (\bm\theta)^T \bm M_2 \bm L_2 \bm T (\bm\theta) \bm L_1^{-1} \big) \bm\Lambda\\ \notag
\bar{\bm h}(\bm q ,\dot{\bm q}, \bm\theta) & \triangleq \bm h_o + \bm h_1 + \bm T(\bm\theta)^T( \bm h_2 + \bm M_2 \bm D \bm\Lambda \dot{\bm x}) \\\label{eq:N_def}
\bm N(\bm\theta) & \triangleq
\begin{bmatrix} \bm I & \bm T(\bm\theta)^T \end{bmatrix}\\ \label{eq:u+}
\bar{\bm u} &  \triangleq \bm N \bm u
\end{align}
\end{subequations}
The reciprocal of \eqref{eq:u+} is given by
\begin{equation} \label{eq:u12}
\bm u = \bm N^+(\bm\theta) \bar{\bm u}    \quad \Leftarrow \quad \min \| \bm u \|,
\end{equation}
where $\bm N^+= \bm N^T(\bm N \bm N^T)^{-1}$ is the {\em
generalized inverse} of matrix $\bm N$, and hence $\bm N \bm N^+ = \bm I$. In view
of \eqref{eq:T} and \eqref{eq:N_def}, the expression of the pseudo-inverse matrix  can be
derived in the following form
\begin{equation} \label{eq:Nplus}
\bm N^+ = \begin{bmatrix} \bm Q^{-1} \\ \bm T \bm Q^{-1} \end{bmatrix} ,
\end{equation}
where the symmetric matrix
\begin{equation}\label{eq:Q}
\bm Q=\bm N \bm N^T = \bm I +\bm T^T \bm T
\end{equation}
is turned out to be a function of only $\bm\rho$ as follow:
\begin{equation}
\bm Q(\bm\rho) = \begin{bmatrix} 2 \bm I &  [\bm\rho \times]  \\
-[\bm\rho \times]  & 2 \bm I - [\bm\rho \times]^2
\end{bmatrix}.
\end{equation}
Therefore, the manipulators' force inputs can be uniquely computed from the input of the reduced order system  \eqref{eq:dyamics_uo} by
\begin{equation} \label{eq:u1u2}
\bm u_1= \bm Q^{-1} \bar{\bm u}, \quad \mbox{and} \quad \bm u_2= \bm T \bm u_1
\end{equation}
Note that relation \eqref{eq:u1u2} requires that approximation of the matrix $\bm Q$ is non-singular. One can observe from \eqref{eq:Q} that matrix $\bm Q$ is the sum of two positive definite matrices and therefore is positive definite and invertible.

Suppose the matrices and vector below are computed from the estimated
parameters, $\hat{\bm \theta}$,
\begin{align*} \notag
\hat{\bm T}  &= \bm T(\hat{\bm \theta}),\\ \hat{\bm Q}& = \bm Q(\hat{\bm \rho}),\\
\hat{\bm M}  &= \bar{\bm M}(\bm q, \hat{\bm\theta}), \\
\hat{\bm h} &= \bar{\bm h}(\bm q,\dot{\bm q}, \hat{\bm\theta}).
\end{align*}
Consider the following inverse dynamics control law
\begin{equation} \label{eq:control}
\bm u := \left\{
\begin{split}
\bm u_1 &=  \hat{\bm Q}^{-1}  \big[ \hat{\bm M} \big(\ddot{\bm x}_d -
\bm G_d \dot{\bm e} - \bm G_p \bm e \big) + \hat{\bm h} \big]\\
\bm u_2 &= \hat{\bm T} \bm u_1,
\end{split} \right.
\end{equation}
where $\bm x_d$ is the desired position trajectory, $\bm e= \bm
x - \bm x_d$ is the position error, and $\bm G_p>0$ and $\bm
G_d>0$ are the feedback gains. Notice that despite the control law \eqref{eq:control} involves  the matrix inversion  computed at the estimated parameters, the matrix is always invertible. In other words, the adaptive controller is {\em singularity-free}. Suppose $\bm
z=[\bm e^T \; \dot{\bm e}^T]^T$ is the state vector error, $\tilde{\bm\theta} =
\bm\theta-\hat{\bm\theta}$ is error in the total parameter set, $\bm G=[\bm G_p \; \bm G_d]$ is the overall gain matrix. Then substituting \eqref{eq:control}
into \eqref{eq:dyamics_uo}, we will arrive at the equation of the
closed-loop system  as
\begin{equation} \label{eq:Az}
\dot{\bm z} = \bm F \bm z + \bm g(\bm z,t)
\end{equation}
where
\begin{align} \notag
\bm F &=\begin{bmatrix} \bm 0 & \bm I \\
-\bm G_p & - \bm G_d \end{bmatrix}, \qquad \mbox{and} \\ \label{eq:d}
\bm g &= \begin{bmatrix} \bm 0 \\
\bar{\bm M}^{-1} \big( (\tilde{\bm N} \hat{\bm
M}- \tilde{\bm M}) \big(\ddot{\bm x}_d - \bm G \bm z \big) - \tilde{\bm h} +
\tilde{\bm N} \hat{\bm h} \big) \end{bmatrix}
\end{align}
is a perturbation term, in which
\begin{align} \notag
\tilde{\bm T} & =\bm T(\bm\theta) - \bm T(\hat{\bm\theta}),\\ \label{eq:tildeN}
\tilde{\bm N} & = \bm N \hat{\bm N}^+ - \bm I  =
\begin{bmatrix} \bm
0 & \tilde{\bm T}^T
\end{bmatrix} \hat{\bm N}^+\\ \notag
\tilde {\bm h} & = \bar{\bm h} - \hat{\bm h} \\ \notag & = \tilde{\bm T}^T (\bm h_2 + \bm M_2 \hat{\bm D} \bm\Lambda \dot{\bm x}) + \bm T^T \bm M_2 \tilde{\bm D} \bm\Lambda \dot{\bm x} , \\ \notag \tilde {\bm M} & =
\bar{\bm M} - \hat{\bm M} \\ \notag & = \big(\tilde{\bm T}^T \bm M_2 \bm L_2 \bm T + \hat{\bm T}^T \bm M_2 \bm L_2 \tilde{\bm T} \big) \bm L_1^{-1} \bm\Lambda.
\end{align}

If the  velocity and acceleration commands are assumed bounded, i.e.,
\begin{equation} \label{eq:cvca}
\norm{\dot{\bm x}_d} \leq c_v \quad
\text{and} \quad \norm{\ddot{\bm x}_d} \leq c_a,
\end{equation}
and the estimators are convergent, then the perturbation
term satisfies the following quadratic growth bound
\begin{equation} \label{eq:perturbation_norm}
\norm{\bm g(\bm z,t)} \leq \big( \kappa_0 + \kappa_1 \norm{\bm z} + \kappa_2 \norm{\bm z}^2\big)
\| \tilde{\bm\theta}(t) \|,
\end{equation}
for some positive constants $\kappa_2$, $\kappa_1$, and $\kappa_0$; see the Appendix~\ref{apdx:alpha_beta}
for detailed derivations. Moreover, for convergent estimator it is reasonable to assume bounded estimation error, i.e.,
\begin{equation} \label{eq:c_theta}
\sup_{\tau \geq 0} \| \tilde{\bm\theta}(\tau) \| = r_{\theta}
\end{equation}
for $r_{\theta}< \infty$. Since a small initial parameter error $\tilde{\bm\theta}(0)$ results in  small parameter error all the time~\cite{Slotine-Li-1991,Aghili-2015b}, there must exist scalar $\alpha>0$ such that $\| \tilde{\bm\theta}(\tau)\| \leq \alpha \| \tilde{\bm\theta}(0)\| $ for all $\tau\geq 0$. The remainder of this section proves that $\bm z(t)$ is {\em ultimately bounded} under the assumption on $r_{\theta}$ being sufficiently small. Furthermore, if $\|\tilde{\bm\theta} \|$ converges to zero, then so does $\| \bm z(t) \|$ for sufficiently large $t$.

Since $\bm F$ is Hurwitz, there exists Lyapunov function
\begin{equation} \label{eq:Lyapunov}
V(\bm z) =  \sqrt{\bm z^T \bm P \bm z}
\end{equation}
with  $\bm P>0$ satisfying
\[ \bm P \bm F + \bm F^T \bm P = -\bm I. \]
Suppose that the domain of analysis is $\bm z \in {\cal D}$, where ${\cal D} = \{\bm z \in \mathbb{R}^6| \norm{\bm z} \leq r_z \}$. Then, the time-derivative of $V$ along trajectories of the
\eqref{eq:Az} satisfies
\begin{align} \notag
\dot V &= \big(- \frac{1}{2}\norm{\bm z}^2 +  \bm z^T \bm P \bm g(\bm z, t) \big)/V \\
\label{eq:dotV} &\leq - \frac{\sigma  \norm{\bm z}^2}{V} +   \frac{ b \norm{\bm z} \| \tilde{\bm \theta} \|}{V}.
\end{align}
where  $\sigma=\frac{1}{2 \bar{\lambda}} - \kappa_1 r_{\theta}$, $b=\bar{\lambda}(\kappa_2 r_z^2 + \kappa_0)$, and
$\overline{\lambda} = \lambda_{\rm max}(\bm P)$. Using the lower and upper bounds of the  Lyapunov function,
\begin{equation} \label{eq:Vbounds}
\sqrt{\underline{\lambda}} \| \bm z \| \leq V \leq \sqrt{\overline{\lambda}} \| \bm z \|,
\end{equation}
where $\underline{\lambda}=\lambda_{\rm min}(\bm P)$, into \eqref{eq:dotV} yields the following Bernoulli differential inequality
\begin{equation}\label{eq:Bernoulli}
\dot V  < - \sigma V + \frac{b}{\sqrt{\underline{\lambda}}} \| \tilde{\bm\theta} \|.
\end{equation}
Here, the coefficient $\sigma$ is positive if
\begin{equation} \label{eq:rtheta}
r_{\theta} < \frac{1}{2\kappa_1 \overline{\lambda}}
\end{equation}
is satisfied. Then, according to the comparison lemma \cite[p. 222]{Khalil-1992,Aghili-2017a}, the solution of \eqref{eq:Bernoulli} must satisfy the inequality
\begin{equation} \notag
V \leq V(0) e^{-\sigma t} + \frac{b}{\sqrt{\underline{\lambda}}} \int_0^t e^{-\sigma(t-\tau)} \|
\tilde{\bm \theta}(\tau) \| d \tau .
\end{equation}
Using the upper and lower limits \eqref{eq:Vbounds} in the above inequality gives
\begin{equation} \label{eq:z(t)}
\| \bm z(t) \| \leq \| \bm z(0) \| \sqrt{\gamma} e^{-\sigma t} +
\frac{b}{\underline{\lambda}} \int_0^t
e^{-\sigma(t-\tau)} \| \tilde{\bm\theta}(\tau) \| d \tau
\end{equation}
where $\gamma= {\overline{\lambda}}/{\underline{\lambda}}$ is the condition number of matrix $\bm P$.
It can be verified that
\begin{equation} \label{eq:rxrt}
\| \bm z(0) \|  \leq  \frac{1}{\sqrt{\gamma}}  r_z \quad \mbox{and} \quad
r_{\theta}  \leq \frac{\sigma \underline{\lambda} }{b} r_z
\end{equation}
ensures the validity of the earlier assumption that $\|\bm z \| \leq r_z$.
Using the expression of $\sigma$ and $b$ in the last inequality in \eqref{eq:rxrt} yields
\begin{equation} \label{eq:rtheta2}
r_{\theta} \leq   \frac{0.5 \underline{\lambda} r_z}{\overline{\lambda}^2 \kappa_0 + \overline{\lambda} \underline{\lambda} \kappa_1 r_z +\overline{\lambda}^2 \kappa_2 r_z^2}
\end{equation}
Notice that restriction \eqref{eq:rtheta2} automatically satisfies \eqref{eq:rtheta} because the RHS of the former inequality is always smaller than that of the latter one. In other words, \eqref{eq:rtheta2} ensures $\sigma>0$. It can be easily verified that
\begin{equation} \notag
r_z = \frac{\kappa_0}{\kappa_2}
\end{equation}
maximize the RHS of \eqref{eq:rtheta2} and therefore
\begin{equation} \label{eq:rtheta3}
r_{\theta} \leq  \alpha \| \tilde{\bm\theta}(0) \| \leq  \frac{1}{2\overline{\lambda} \big(  \kappa_1 + \gamma(\kappa_0 +\kappa_2) \big)}
\end{equation}
is the largest bound on the parameter estimation error that still guarantees ultimate boundedness of $\bm z(t)$. In this case, it is apparent from the response of the force system \eqref{eq:z(t)}, that $\| \bm z(t) \|$ converges to zero if the input signal $\| \tilde{\bm\theta}(t) \|$ does so. In other words, restrictions \eqref{eq:cvca}, \eqref{eq:rtheta3}  and $\| \bm z(0) \| \leq \kappa_0 / ( \kappa_2 \sqrt{\gamma})$ ensure that the control error $\bm z(t)$ is ultimate bounded. Furthermore, if $\|\tilde{\bm\theta}\|\rightarrow 0$ as $t\rightarrow \infty$ then  $\|\bm z(t)\|\rightarrow 0$ as $t\rightarrow \infty$.

The above development can be summarized in the following.
Assume the following assumptions
\begin{enumerate}
\item The direction of the angular velocity vector does not remain constant over time, i.e., convergence of the estimators according to Propositions \ref{prop:eta} and \ref{prop:rho}.
\item The velocity and acceleration command signals are bounded
\item The initial kinematic parameter error is sufficiently small, i.e., condition \eqref{eq:rtheta3} is satisfied
\end{enumerate}
Then, the convergence and stability of the estimation and control process are guaranteed.

\section{Conclusions}
Adaptive self-tuning control of cooperative manipulators carrying a common
object in the presence of position/orientation uncertainties in the
closed-kinematic loop was presented. This allows accurate motion tracking performance in the presence of
uncertainties in the relative position/orinetaiton of the manipulators' bases as well as those of their end-effectors while grasping the object. The advantage of the control scheme is that it does not use either the measurement of the contact forces to compensate for the geometric uncertainties or a high precision end-point sensing device for fine calibration. However, the practical limitations of the method are that it assumes $i)$ a rigid-body common object and $ii)$ an initial coarse estimate of the kinematic parameters. Two cascaded recursive least-squares estimators were used to estimate the kinematic parameters in real-time. A cooperated controller was developed to take the
estimated parameters in order to achieve motion tracking of a reference trajectory while minimizing the weighted Euclidean norm of the actuator forces. The proposed adaptive controller is singularity-free because
it involves inverting a matrix computed at the estimated parameters that turned out to be always invertible.
The convergence and stability of the combined system of the estimator and controller have been thoroughly analyzed and the results has shown  that tracking performance can be achieved provided that $i)$ the direction of the angular velocity of the object changes over time; $ii)$ the initial value error of the kinematic parameters is sufficiently small.

%-----------------------------------------------------------
\section*{Appendix A} \label{apdx:least-squares}
%-----------------------------------------------------------
Consider the following linear parametrization form
\begin{equation}
\bm y_k = \bm W_k \bm a_k + \bm e_k.
\end{equation}
The least-squares method finds the current parameter estimate $\hat{\bm a}_k$ which minimizes the cost function
\begin{equation}
J = \sum_{i=1}^k  e^{-\mu(t_k -t_i)}\| (\bm y_i - \bm W_i \hat{\bm a}_k) \|^2
\end{equation}
The parameter update law is
\begin{equation}
\hat{\bm a}_{k+1} = \hat{\bm a}_k + \bm K_k(\bm y_k - \hat{\bm y}_k)
\end{equation}
where $\hat{\bm y}_k = \bm W_k \hat{\bm a}_k$, $\varrho=e^{-\mu h}$, and
\begin{align*}
\bm K_k &= \bm P_k \bm W_k^T \big(\varrho \bm I + \bm W_k \bm P_k \bm W_k^T )^{-1} \\
\bm P_{k+1} & =(\bm I - \bm K_k \bm W_k) \bm P_k/ \varrho
\end{align*}
are the gain and covariance matrices.

%-----------------------------------------------------------
\section*{Appendix B} \label{apdx:cartesian_dynamics}
%-----------------------------------------------------------
Dynamics equations of every manipulator in the joint space is
described by
\begin{equation} \label{eq:Mrddotq}
{\bm M}'_i \ddot{\bm q}_i + {\bm h}'_i(\bm q_i, \dot{\bm  q}_i) = \bm\tau_i
+ \bm J_i^T \bm f_i
\end{equation}
Substituting the joint acceleration from $\ddot{\bm q}_i=\bm J_i^{-1}
\ddot{\bm x}_i - \bm J_i^{-1} \dot{\bm J}_i\dot{\bm q}_i$ into
\eqref{eq:Mrddotq} and then multiply the resultant equation by $\bm
J_i^{-T}$ yields \eqref{eq:dynamics_task}, in which the dynamics parameters in the task space are related to those in the joint space by
\begin{align*}
\bm M_i & \triangleq \bm J_i^{-T} {\bm M}'_i \bm J_i^{-1} \\
\bm h_i &\triangleq \bm J_i^{-T} {\bm h}'_i - \bm M_i \dot{\bm J}_i
\dot{\bm q}
\end{align*}
%-----------------------------------------------------------
\section*{Appendix C} \label{apdx:alpha_beta}
%-----------------------------------------------------------
It is known the mass matrix is a positive definite matrix with lower and upper bounded limited norms such that the following inequality holds for $c_m, c_M>0$
\begin{equation} \label{eq:Mi_bound}
c_m \bm I \leq \bm M_i(\bm q) \leq c_M \bm I \qquad \forall i=1,2.
\end{equation}
The nonlinear vector $\bm h(\bm q, \dot{\bm q})$ contains the gravitational plus the Coriolis and centrifugal terms. For revolute joint robots, the function has a periodic dependence  of $\bm q$ and quadratic dependence on $\dot{\bm q}$~\cite[p.144]{Canudas-Siciliano-Bastin-book-1996}. Therefore, the nonlinear vector can be assumed bounded as
\begin{align} \notag
\norm{\bm h_i(\bm q, \dot{\bm q})}  &\leq c_g + c_h
\norm{\dot{\bm x}}^2 \\
\label{eq:normhi}& \leq c_g+ c_h c_v + c_h \norm{\bm z}^2.
\end{align}
Similar to \eqref{eq:Mi_bound} and \eqref{eq:normhi}, the bound limits  can be expressed for the interconnected system \eqref{eq:dyamics_uo} as
\begin{equation} \label{eq:M_bound}
\bar c_m \bm I \leq \bar{\bm M}(\bm q) \leq \bar c_M \bm I,
\end{equation}
and
\begin{equation}\label{eq:normh}
\norm{\bar{\bm h}(\bm q, \dot{\bm q})} \leq  \bar c_g+ \bar c_h c_v + \bar c_h \norm{\bm z}^2.
\end{equation}
The inverse matrix $\hat{\bm N}^+$ always exists and hence
the  matrix must be  upper bounded, i.e.,
\begin{equation} \label{eq:N_bound}
\hat{\bm N}^+  \leq c_n \bm I \qquad \forall  \bm\theta \in \Theta
\end{equation}
for $c_n>0$. Moreover, since $\bm T$ and $\bm D$ are Lipschitz in
terms of the parameter variables, we can say there exist positive scalar $\varepsilon_{t}$ and $\varepsilon_{d}$ such that
\begin{align} \label{eq:T_lip}
\| \tilde {\bm T} \| &=  \| \bm T(\bm\theta) - \bm T(\hat{\bm \theta}) \|  \leq \varepsilon_{t} \|
\tilde{\bm\theta} \| \\
\| \tilde {\bm D} \| &=  \| \bm D(\bm\theta) - \bm D(\hat{\bm \theta}) \|  \leq \varepsilon_{d} \|
\tilde{\bm\theta} \|
\end{align}
From inequalities \eqref{eq:N_bound} and \eqref{eq:T_lip} and
identity \eqref{eq:tildeN} we  get
\begin{equation} \label{eq:NM}
\| \tilde{\bm N} \|  \leq c_t c_n \| \tilde{\bm\theta} \|.
\end{equation}
Using the norm properties and the above inequalities, we arrive at
the following inequality
\begin{align} \notag
& \| \tilde{\bm h} \|  \leq  \big(\varepsilon_t c_g + \varepsilon_t c_h c_v + (\varepsilon_t + \varepsilon_d) c_M c_d c_{\lambda} c_v  +   (\varepsilon_t + \varepsilon_d) c_M c_d c_{\lambda}\| \bm z \| + \varepsilon_t c_h \| \bm z \|^2 \big) \|
\tilde{\bm \theta} \| \\ \notag
& \| \tilde {\bm N} \hat{\bm M}  - \tilde{\bm M} \|  \leq  \varepsilon_t c_M ( c_n  + 2 c_l c_{\lambda}c_t ) \| \tilde{\bm\theta} \|  \\ \label{eq:hMNh}
& \| \tilde{\bm N} \hat{\bm h} \|  \leq  \varepsilon_t  c_n (\bar c_g + \bar c_h c_v \| \bm z \| + \bar c_h \| \bm z \|^2 ) \| \tilde{\bm\theta} \|
\end{align}
where $c_t = \| \bm T \|$   and $c_d = \| \bm D \|$ for all $\bm\theta\in \Theta$, $c_{\lambda} = \| \bm\Lambda \|$, and $c_l  = \bar{\sigma}(\bm L_2) / \underline{\sigma}(\bm L_1)$. Finally using bound limits \eqref{eq:NM} in the expression of the
perturbation term \eqref{eq:d} yields inequality
\eqref{eq:perturbation_norm}, in which
\begin{align*}
\kappa_0 =& (\varepsilon_t c_M c_a(c_n + 2 c_l c_{\lambda} c_t) + \varepsilon_t c_g + \varepsilon_t c_h c_v +(\varepsilon_t + \varepsilon_d) c_M c_d c_{\lambda} c_v+ \varepsilon_t c_n \bar c_g)/c_m\\
\kappa_1  =& (\varepsilon_t c_M(c_n + 2 c_l c_{\lambda} c_t)\| \bm G \|+ (\varepsilon_t + \varepsilon_d )c_M c_d c_{\lambda} + \varepsilon_t c_n \bar c_h c_v )/c_m\\
\kappa_2  =& (c_h + c_n \bar c_h) \varepsilon_t /c_m
\end{align*}

\section*{Acknowledgement}
This work was supported by  Natural Sciences and Engineering Research Council of Canada (NSERC) under grant RGPIN/288255-2011.

%-------------------------------------------------------------
\bibliographystyle{IEEEtran}
%\bibliography{references}
%\end{document}
%-------------------------------------------------------------

\end{document}